\PassOptionsToPackage{table}{xcolor}
\documentclass{article}
\usepackage[T1]{fontenc}
\usepackage{iclr2027_conference,times}

\usepackage{amsmath,amsfonts,bm}

\def\eqref#1{equation~\ref{#1}}

\def\1{\bm{1}}

\DeclareMathAlphabet{\mathsfit}{\encodingdefault}{\sfdefault}{m}{sl}
\SetMathAlphabet{\mathsfit}{bold}{\encodingdefault}{\sfdefault}{bx}{n}

\usepackage{graphicx}
\usepackage{wrapfig}
\usepackage{url}
\usepackage{xcolor}
\usepackage{booktabs}
\usepackage{hyperref}
\hypersetup{colorlinks=true,linkcolor=blue,citecolor=blue,urlcolor=blue,bookmarksopen=true}
\newcommand{\paperref}[2]{\hyperref[#2]{#1~\ref*{#2}}}
\newcommand{\panelref}[2]{\hyperref[#1]{Figure~\ref*{#1}(#2)}}
\title{Efficient Scene-Text Refinement\\via Pixel-Space Grouped Image Generation}
\author{Yaoli Liu\textsuperscript{1,2,\ensuremath{\dagger}}, Guo-Hua Wang\textsuperscript{2}\thanks{Corresponding authors.}, Yiyang Chen\textsuperscript{3,2,\ensuremath{\dagger}}, Qing-Guo Chen\textsuperscript{2},\\
\bfseries Zhao Xu\textsuperscript{2}, Weihua Luo\textsuperscript{2}, Yao-Xiang Ding\textsuperscript{1,\textasteriskcentered}\\[4pt]
\normalfont\textsuperscript{1} State Key Lab of CAD\&CG, Zhejiang University (\url{zju.edu.cn})\\
\normalfont\textsuperscript{2} Alibaba Group (\url{alibaba-inc.com})\\
\normalfont\textsuperscript{3} State Key Laboratory of General Artificial Intelligence, Peking University (\url{pku.edu.cn})\\[3pt]
\normalfont\small\texttt{yaoliliu8@gmail.com}\quad\texttt{wangguohua@alibaba-inc.com}}
\iclrfinalcopy
\begin{document}
\maketitle
\begingroup
\renewcommand{\thefootnote}{\ensuremath{\dagger}}
\footnotetext{Work done during the internship at Alibaba Cloud Computing.}
\endgroup
\lhead{\footnotesize Efficient Scene-Text Refinement via Pixel-Space Grouped Image Generation}
\begin{abstract}
Generated images often have coherent layouts and visual content, yet local errors in small text and complex characters still limit their readability and practical use. Scene-text refinement requires \emph{accuracy}, \emph{preservation}, and \emph{efficiency}. Existing methods based on variational autoencoders (VAEs) can lose fine strokes during reconstruction, while full-image editing can shift background colors. In this work, we introduce \emph{PixText}, a model that formulates local text refinement as grouped image generation in pixel space. Conditioned on the original image, PixText generates only the target regions, preserves unedited content, and uses MeanFlow for four-step generation. We further introduce \emph{Text Refinement Bench} to evaluate text correction and image preservation in generated images. On this benchmark, PixText achieves 83.06\% glyph pass and 94.66\% exact target-character OCR accuracy, versus 36.77\% and 52.31\% for the strongest baseline. Its four-step MeanFlow variant retains an 81.84\% pass rate with less than one GPU-second of amortized generation cost per image.
\end{abstract}

\section{Introduction}
Large text-to-image models have made rapid progress in text rendering and layout. Images often exhibit reasonable layouts and mostly correct text, yet errors persist in small text and complex characters. These errors occupy only a small part of an image but directly affect its readability and practical usability.

Further improving text accuracy within a general-purpose generative model requires balancing text fidelity against overall visual quality. For example, the pretraining experience of LongCat-Image shows that synthetic text data improves text accuracy but can impair overall visual coherence, motivating dynamic adjustment of its sampling proportion and its removal toward the end of pretraining \citep{team2025longcat}. Post-training for text accuracy faces a similar challenge: excessive optimization of text rewards can make text disproportionately large or prominent, compromising aesthetics and visual coherence \citep{diao2026sensenova,team2026firered,mao2026image,jin2025inference,chen2026mapreduce,kang2026transferability}. These observations motivate \emph{separating text refinement from overall image generation}: preserving the layout, style, and visual content produced by the base model while precisely correcting only the erroneous text regions.

Scene-text refinement calls for \emph{accuracy, preservation, and efficiency}: correcting text faithfully, leaving unedited content unchanged, and keeping computational cost low. Existing text generation and editing methods, however, typically redraw entire text regions and can alter their original layout, style, and typeface. Most build on large latent-space generative models \citep{tuo2024anytext,tuo2024anytext2,lan2025fluxtext,xie2026textflux}, whose ability to generate and refine small text is constrained by variational autoencoder (VAE) reconstruction \citep{yu2026pixeldit}, while also incurring the cost of a large generative network and VAE encoding and decoding. Applying these methods directly to localized corrections of a few characters therefore makes it difficult to satisfy all three requirements.

To tackle these challenges, we introduce \emph{PixText}, a text generation and refinement model based on grouped image generation in pixel space. First, we model images directly in pixel space, removing the VAE reconstruction bottleneck for small-character strokes as well as the texture degradation and color shifts introduced by local encoding and decoding. Second, we use the original image as the source that provides context and organize the regions to be refined as a group of image elements. The model predicts only the item pixels, which are then pasted back into the original image. By avoiding full-image regeneration, this design preserves unedited regions and reduces the background color drift and paste seams introduced by full-image generation. Finally, four-step MeanFlow \citep{geng2026mean} reduces amortized generation cost to less than one GPU-second per image.

Meanwhile, we propose simple, effective training and inference improvements. We identify the influence of the training timestep distribution on patch-boundary artifacts and mitigate them with uniform sampling, observing consistent improvements in our model and MiniT2I \citep{minit2i2026}. We also revise the $x$-prediction sampling update to reduce speckle from incomplete denoising.

As a final contribution, we propose a new benchmark for text refinement. The existing AnyText Bench \citep{tuo2024anytext} primarily evaluates conventional text generation and lacks a dedicated evaluation of localized text errors in generated images. We therefore construct \emph{Text Refinement Bench} from model-generated images, providing character-level error annotations, correction targets, and a unified evaluation protocol. The benchmark evaluates correction accuracy, background preservation, and inference efficiency. Experiments show that PixText outperforms existing baselines on text refinement and achieves performance close to or above the best existing methods on AnyText Bench at lower computational cost.

\section{Related Work}
\subsection{Scene-Text Generation and Editing}
Early scene-text editing methods modeled content and style to replace text while preserving its appearance \citep{wu2019srnet,roy2020stefann,yang2020swaptext,krishnan2023textstylebrush,lee2021rewritenet,qu2023mostel}. Diffusion models improved text generation and editing through layout, glyph, and character conditioning \citep{chen2023textdiffuser,chen2024textdiffuser2,ma2023glyphdraw,ma2025glyphdraw2,yang2023glyphcontrol,liu2024glyphbyt5,liu2024glyphbyt5v2,tuo2024anytext,tuo2024anytext2,ji2023iffSTE,chen2023diffute,zhao2024udifftext,wang2025dreamtext}. Subsequent work advanced glyph accuracy, style control, and multilingual editing \citep{zeng2024textctrl,wang2025glyphmastero,yan2025textmaster,fang2025rs-ste,lan2025fluxtext,xie2026textflux,chen2026styletextgen}, extending to open-vocabulary, training-free, and instruction-driven unified text editing \citep{li2026self,gunawan2026omnitext,li2026towards,wang2026textwand,ma2026um-text,zhang2026weedit,lin2026textsculptor,wang2026textrefine}, as well as text correction in generated images \citep{shimoda2025type}. We focus on refining a few erroneous characters in generated images: jointly generating multiple target regions directly in pixel space avoids the effects of VAE compression and reconstruction on fine strokes while preserving unedited content. Four-step MeanFlow improves efficiency while keeping the amortized generation cost below one GPU-second per image.
\subsection{Pixel-Space Generation}
Pixel-space generation models images directly, avoiding information loss from latent compression. Early work explored high-resolution modeling and computational efficiency \citep{hoogeboom2023simplediffusion,hoogeboom2025simplerdiffusion,jabri2022RIN,crowson2024scalable,atzmon2024edify}, while subsequent methods improved quality and efficiency through global--local modeling and frequency decomposition \citep{chen2025pixelflow,wang2026pixnerd,yu2026pixeldit,chen2026dip,ma2026deco,lin2026frepix}. JiT demonstrated the effectiveness of predicting clean images directly \citep{li2026back}; related studies explored perceptual supervision, generative trajectories, and text-to-image generation and editing \citep{ma2026pixelgen,baade2026latent,minit2i2026,cai2026hidream}, while transferring pretrained latent-space models reduced training costs \citep{chen2026l2p,chen2026asymmetric,jiang2026empirical}. Meanwhile, MeanFlow and its successors model average velocity for few-step generation, extending this approach to pixel space \citep{geng2026mean,geng2026improved,lu2026one}. Building on these advances, we focus on local text refinement conditioned on the original image, jointly generating the target regions to combine glyph accuracy, content preservation, and efficiency.

\section{Method}
\begin{figure}[t]
  \centering
  \includegraphics[width=\linewidth]{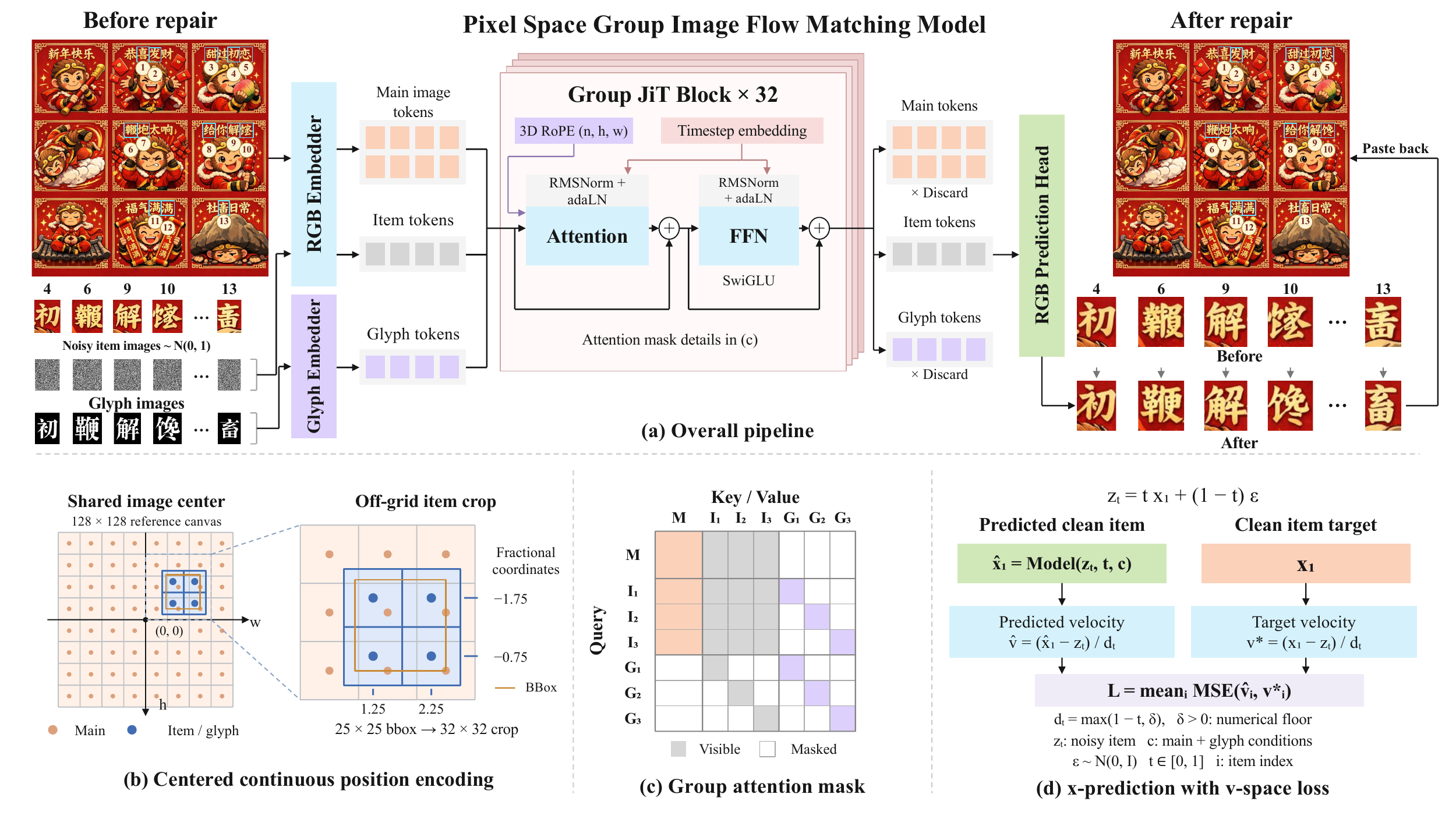}
  \caption{Overview of PixText. A main image provides context, noisy item images represent the regions to be generated, and glyph images specify the target text. Position encoding aligns each item and its glyph with the main image. Group JiT blocks predict the item pixels, which are composed into the original image.}
  \label{fig:pipeline}
\end{figure}

\subsection{Pixel-Space Generation for Quality and Efficiency}
\label{sec:pixel-space}

Latent-space text generation and editing methods rely on VAE compression and reconstruction \citep{li2026self,chen2023textdiffuser,zeng2024textctrl,tuo2024anytext,tuo2024anytext2,lan2025fluxtext,xie2026textflux}, which can lose fine strokes and incur additional decoding cost. Related work on autoregressive models also identifies visual-tokenizer reconstruction as a bottleneck for text detail and improves glyph quality by enhancing the decoder \citep{mao2026residual}. In \paperref{Figure}{fig:vae-reconstruction}, a character approximately 20 pixels high becomes difficult to recognize after reconstruction with Stable Diffusion, FLUX.1, and Qwen-Image VAEs \citep{rombach2022high,labs2025flux,wu2025qwen}. For edits involving only a few characters, decoding only the target region may lower VAE decoding cost, but the smaller region gives the decoder less image context. \paperref{Figure}{fig:vae-tile} compares full-image and local FLUX.1-VAE decoding from the same latent representation obtained by encoding the full image: the independently decoded region exhibits a visible color mismatch when pasted back, even with 32 pixels of additional context.

\begin{figure}[t]
  \centering
  \includegraphics[width=\linewidth]{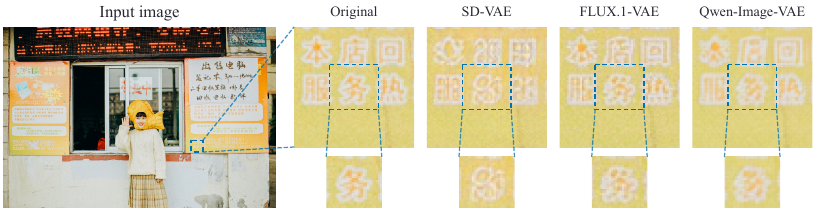}
  \caption{Small-character degradation during VAE reconstruction. Each VAE uses full-image encoding and posterior-mode decoding in FP32.}
  \label{fig:vae-reconstruction}
\end{figure}

\begin{figure}[t]
  \centering
  \includegraphics[width=\linewidth]{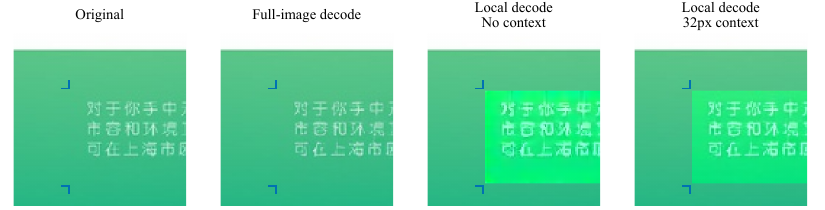}
  \caption{Local FLUX.1-VAE decoding causes visible color shifts when pasted back into the original image, even with 32 pixels of context.}
  \label{fig:vae-tile}
\end{figure}

We therefore build a flow matching model directly in pixel space, eliminating VAE encoding and decoding, and adopt the simple JiT architecture \citep{li2026back}. \paperref{Table}{tab:vae-cost} shows that, under matched hardware and resolution, a single call to the JiT-L/16 output head takes less than 1\% of the decoding time of the evaluated autoencoders, which also include FLUX.2 and DC-AE \citep{flux-2-2025,chen2025deep}. Among pixel-space models, its output head is also lighter than those of PixelDiT, DiP, and DeCo \citep{yu2026pixeldit,chen2026dip,ma2026deco}. These heads run at every denoising step, so their costs accumulate across sampling steps, motivating us to retain JiT's lightweight design.

\begin{table}[t]
\centering
\caption{Decoder and pixel output head costs at $1024\times1024$ resolution (single NVIDIA H800, FP16, batch size 1). The best and second-best results in each row are bolded and underlined, respectively.}
\label{tab:vae-cost}
\begingroup
\definecolor{pixtextsky}{HTML}{DDF3FC}
\footnotesize
\setlength{\tabcolsep}{2pt}
\renewcommand{\arraystretch}{1.25}
\resizebox{\linewidth}{!}{%
\begin{tabular}{>{\columncolor{gray!20}}l||>{\columncolor{white}}r>{\columncolor{gray!10}}r>{\columncolor{white}}r>{\columncolor{gray!10}}r>{\columncolor{white}}r|>{\columncolor{gray!10}}r>{\columncolor{white}}r>{\columncolor{gray!10}}r|>{\columncolor{pixtextsky}}r}
\hline\hline
Module & \shortstack{SD\\VAE} & \shortstack{FLUX.1\\VAE} & \hspace{3pt}\shortstack{\strut Qwen-Image\\\strut VAE}\hspace{3pt} & \shortstack{FLUX.2\\VAE} & \shortstack{DC-AE\\f32c32} & \shortstack{DiP\\XL/16} & \shortstack{PixelDiT\\1300M} & \shortstack{DeCo\\XXL/16} & \shortstack{JiT\\L/16} \\
\hline\hline
Params (M) $\downarrow$ & 49.49 & 49.55 & 73.30 & 49.62 & 159.22 & \underline{6.97} & 105.06 & 12.61 & \textbf{2.89} \\
Latency (ms) $\downarrow$ & 90.89 & 90.54 & 69.94 & 90.76 & 55.38 & \underline{5.65} & 15.01 & 8.18 & \textbf{0.12} \\
GFLOPs $\downarrow$ & 10470.4 & 10472.2 & 11682.6 & 10474.7 & 8368.6 & 513.5 & 868.9 & \underline{142.1} & \textbf{6.4} \\
\hline\hline
\end{tabular}}
\endgroup
\end{table}

For the prediction target, we follow the low-dimensional manifold perspective of JiT, HiDream-O1, and miniT2I \citep{li2026back,cai2026hidream,minit2i2026} and use $x$-prediction to predict the clean image directly. We then convert the prediction to velocity for computing the training loss and performing sampling updates.
Let $x_1$ denote a clean image and $\epsilon\sim\mathcal{N}(0,I)$ Gaussian noise. We construct a noisy input along a linear path from noise to data:
\begin{equation}
 z_t=t x_1+(1-t)\epsilon,\qquad t\in[0,1].
 \label{eq:pixel-flow-path}
\end{equation}
The model directly predicts the clean image as $\hat{x}_1=f_\theta(z_t,t,c)$, where $c$ denotes the conditioning input. As illustrated in \panelref{fig:pipeline}{d}, we convert the prediction and target to velocity space to compute the training loss:
\begin{equation}
 \hat{v}_t=\frac{\hat{x}_1-z_t}{d_t},\qquad
 v_t^*=\frac{x_1-z_t}{d_t},\qquad
 d_t=\max(1-t,\delta),
 \label{eq:pixel-flow-velocity}
\end{equation}
where $\delta>0$ is a numerical floor. We minimize the mean squared error between predicted and target velocities.

\subsection{Text Editing and Refinement as Grouped Image Generation}
\begin{figure}[t]
  \centering
  \includegraphics[width=\linewidth]{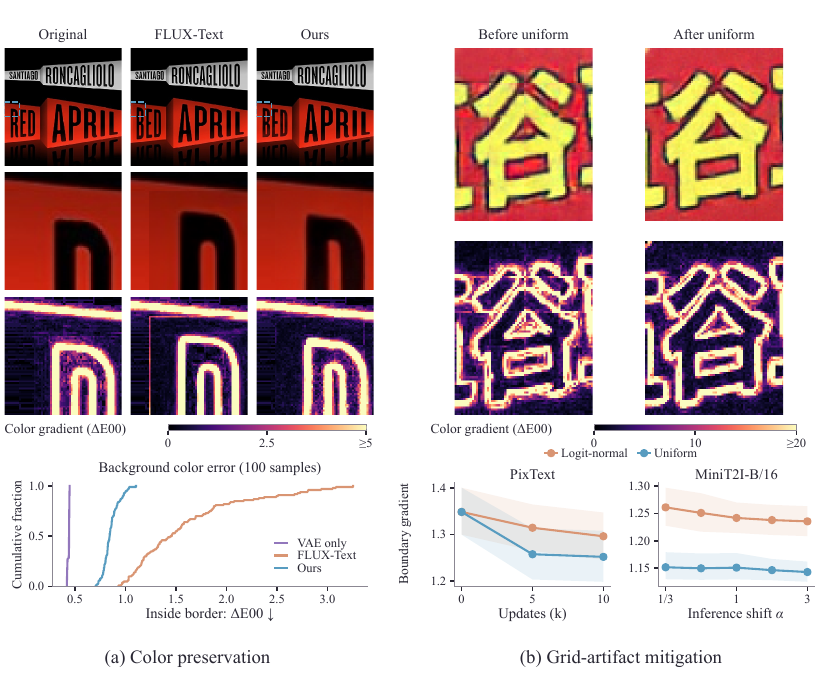}
  \caption{Color preservation and grid-artifact mitigation. (a) Full images, local close-ups, and color-gradient maps, with background color-error distributions. (b) Generated images and color-gradient maps before and after uniform sampling, with PixText and MiniT2I boundary-gradient curves. Heatmaps share a scale within each panel; shading denotes 95\% bootstrap intervals.}
  \label{fig:color-drift}
  \label{fig:uniform-sampling}
\end{figure}
Recent scene-text editing methods \citep{lan2025fluxtext,xie2026textflux,li2026self} use full-image inpainting: even for local text edits, they update and decode full-image latents, reconstructing background that requires no modification. This full-image generation can shift background colors, leaving visible seams after crop-and-paste editing (\panelref{fig:color-drift}{a}). The shift exceeds the VAE-only reconstruction error, showing that pasting alone does not ensure color consistency between the edited region and the original image.

We formulate text editing and refinement as grouped image generation in pixel space, separating context from generation targets. During both training and inference, we erase the existing text in each target region through local blurring before using the source image as the main-image condition. The resulting image serves as a \emph{main image} providing context, while each region to be edited forms an \emph{item image}. The model predicts only the item pixels and pastes them into the original. The extent of generation therefore follows the size of the edit regions, while pixels outside them are retained directly. We denote the $N$ clean items by $x_1=\{x_1^i\}_{i=1}^{N}$ and their noisy counterparts by $z_t=\{z_t^i\}_{i=1}^{N}$. Conditioned on the main image and target glyph images, denoted by $c$, the model jointly predicts $\hat{x}_1=f_\theta(z_t,t,c)$.

The input comprises main-image, noisy item-image, and glyph-image tokens, which provide scene context, represent generation targets, and specify text content, respectively. At sampling time, item images are initialized from $\mathcal{N}(0,I)$. Main and item images share an RGB embedder. Glyph images are rendered with PIL in single-channel grayscale and mapped to the same feature dimension through a separate glyph embedder. Target text is specified entirely through these glyph images, removing the need to deploy an additional text encoder.

As illustrated in \panelref{fig:pipeline}{b}, we use \emph{Centered Continuous Position Encoding} (CCPE) to align the main image and independently cropped items across scales in the coordinate system of the original image. Let the original image have size $H\times W$, the patch size be $p$, and $(y,x)$ denote a patch center in original-image pixel coordinates. Its 3D RoPE coordinates \citep{su2024roformer} are
\begin{equation}
 \boldsymbol{\pi}=\left(n,\frac{y-H/2}{p},\frac{x-W/2}{p}\right),
 \qquad n=\begin{cases}
 0,&\text{main image},\\
 1,&\text{item image},\\
 2,&\text{glyph image}.
 \end{cases}
 \label{eq:centered-continuous-position}
\end{equation}
Each glyph shares the spatial coordinates of its paired item and is distinguished by the role coordinate $n$. Continuous spatial coordinates allow item crops to start at arbitrary pixel locations without aligning to the main-image patch grid. Only crop dimensions need to be expanded to multiples of $p$, avoiding extra padding from grid alignment. Across 116,038 text regions, this reduces the number of item tokens by 13.7\% relative to grid-aligned crops; \paperref{Appendix}{app:ccpe-tokens} provides the comparison.

We adapt JiT blocks for grouped image generation and call the resulting blocks \emph{Group JiT Blocks}. These tokens are jointly processed by 32 such blocks. As illustrated in \panelref{fig:pipeline}{c}, the main image and items interact bidirectionally, and different items attend to one another to share scene information and editing context. Cross-image attention involving each glyph is restricted to bidirectional interaction with its paired item, preserving the correspondence between text conditions and editing targets. We inject timestep conditions through adaLN-Zero \citep{peebles2023scalable}: noisy items receive the current timestep $t$, while the main image and glyphs use the fixed clean endpoint $t=1$, distinguishing generation targets from known conditions. We use FlexAttention \citep{dong2024flex} to batch variable-length image groups; implementation details are provided in \paperref{Appendix}{app:flex}.

\subsection{Training and Sampling}

We train PixText directly for local text editing, conditioning on the main image and glyphs while adding noise to and predicting only the item images. We choose the editing granularity according to the writing system, using character-level regions for Chinese and word-level regions for alphabetic writing systems such as English to support fine-grained refinement. Glyph conditions are rendered from the recognized text. Training uses the noise path and velocity-space objective in \paperref{Section}{sec:pixel-space}. We average the loss over pixels and channels within each item, then equally over items:
\begin{equation}
 \mathcal{L}=\mathbb{E}_{x_1,c,t,\epsilon}\!\left[
 \frac{1}{N}\sum_{i=1}^{N}
 \frac{\|\hat{v}_t^i-v_t^{*,i}\|_2^2}{D_i}
 \right],
 \label{eq:pixel-flow-loss}
\end{equation}
where $D_i$ is the number of scalar pixel values in item $i$.


Grid artifacts aligned with patch boundaries have been observed across JiT-based pixel-space models \citep{minit2i2026,jiang2026empirical,he2026bend}. We find that these artifacts can be mitigated by a simple change to the training timestep distribution. After initial training with logit-normal sampling, we switch to uniform sampling and continue training. Generated images and color-gradient maps in \panelref{fig:uniform-sampling}{b} show weaker patch-boundary discontinuities after uniform sampling; the bottom-left plot compares both training strategies from the same initialization. Uniform sampling yields lower patch-boundary gradients at both continuation checkpoints, indicating weaker grid artifacts. We further evaluate this strategy on MiniT2I-B/16. Starting from the same pretrained checkpoint used for the official logit-normal fine-tuning, we adopt uniform timestep sampling during post-training without modifying the architecture. The bottom-right plot in \panelref{fig:uniform-sampling}{b} shows lower boundary gradients than the official logit-normal fine-tuned model across all five inference schedules, indicating that this simple post-training adjustment also mitigates grid artifacts in another pixel-space model. \paperref{Appendix}{app:training-sampling} provides the experimental details.

The sampling implementations of JiT and MiniT2I impose a lower bound on $1-t$ when converting clean-image predictions into velocities, preventing the denominator from approaching zero near the denoising endpoint \citep{li2026back,minit2i2026}. We observe that this floor can weaken the final updates and leave residual speckle. We instead formulate the update directly in terms of the clean prediction, so the final step returns it in full. We hold the main-image and glyph conditions $c$ fixed and initialize the items from Gaussian noise, $z_0^i\sim\mathcal{N}(0,I)$. For a time sequence $0=t_0<\cdots<t_K=1$, the Euler update is
\begin{equation}
 z_{t_{k+1}}^i = z_{t_k}^i+\frac{t_{k+1}-t_k}{1-t_k}\left(\hat{x}_{1,k}^i-z_{t_k}^i\right),\qquad \hat{x}_{1,k}=f_\theta(z_{t_k},t_k,c).
 \label{eq:pixel-flow-sampling}
\end{equation}
When the final step reaches $t_{k+1}=1$, this update directly returns the clean-image prediction $\hat{x}_{1,k}^i$. \paperref{Figure}{fig:sampling-endpoint} shows reduced background speckle when only the final update is replaced. Additional matched examples and sampling settings are provided in \paperref{Appendix}{app:training-sampling}.
\begin{figure}[t]
  \centering
  \includegraphics[width=\linewidth]{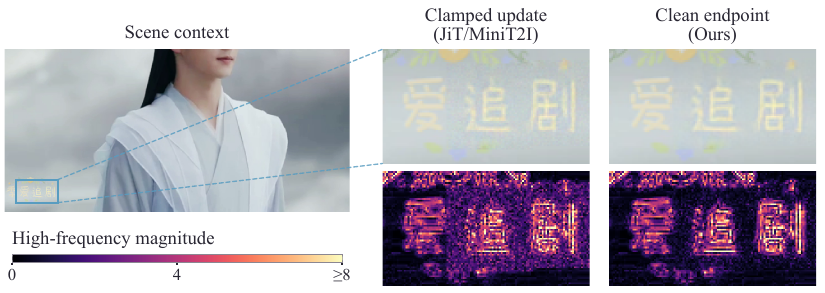}
  \caption{Effect of the final sampling update on residual speckle. Left: scene context. Right: matched close-ups and high-frequency maps show reduced background fluctuations when returning the clean prediction. Both outputs differ only in the final update along a shared trajectory.}
  \label{fig:sampling-endpoint}
\end{figure}

\subsection{Efficient Generation with MeanFlow}
To reduce the number of sampling steps, we distill a trained PixText teacher using MeanFlow to learn the average velocity over a time interval \citep{geng2026mean}. The student is initialized from the teacher and additionally conditioned on the interval endpoint. Let $z_t$ collect all items to be generated at time $t$, and let $v_\phi(z_t,t,c)$ denote the frozen teacher's instantaneous velocity. Along a teacher trajectory, the average velocity over $[t,s]$, with $t<s$, satisfies the MeanFlow identity:
\begin{equation}
 u(z_t,t,s,c)=\frac{z_s-z_t}{s-t}=v_\phi(z_t,t,c)+(s-t)\mathcal{D}_t u(z_t,t,s,c),
 \label{eq:meanflow-identity}
\end{equation}
where $\mathcal{D}_t u=\partial_t u+(\partial u/\partial z_t)v_\phi$ is the total derivative along the teacher velocity, holding $s$ and $c$ fixed. This identity connects interval displacement to the teacher velocity at the current state, providing supervision without explicitly integrating the teacher trajectory.

We independently sample two uniform timesteps and sort them to obtain $t<s$, constructing training states with the noise path defined above. Following AnyFlow, we use the Differential Derivation Equation (DDE) approach to estimate this total derivative by finite differences \citep{gu2026anyflow}. Let $v_t=v_\phi(z_t,t,c)$ be the instantaneous velocity predicted by the frozen teacher at the current state. We perturb the state along this fixed teacher velocity together with the starting time, holding the endpoint $s$ and condition $c$ fixed:
\begin{equation}
 U_\theta(\delta)=u_\theta(z_t+\delta v_t,t+\delta,s,c),\qquad
 \widehat{\mathcal{D}}_t u_\theta(z_t,t,s,c)
 =\frac{U_\theta(\delta_+)-U_\theta(\delta_-)}{\delta_+-\delta_-}.
 \label{eq:meanflow-dde}
\end{equation}
Here $\delta_-\leq 0\leq\delta_+$ and $\delta_-<\delta_+$. We normally use $(\delta_+,\delta_-)=(h,-h)$, giving a central difference with step size $h>0$. Near boundaries, we use the one-sided pair $(h,0)$ or $(0,-h)$ and reduce $h$ as needed to keep the perturbed starting times within $[0,s]$ and on the same branch of the denominator clamp used in velocity conversion. The finite-difference predictions are evaluated without gradient tracking. The distillation objective is
\begin{equation}
 \mathcal{L}_{\mathrm{MF}}=
 \mathbb{E}\!\left[
 \frac{1}{D}\left\|
 u_\theta(z_t,t,s,c)-\operatorname{sg}\!\left[
 v_\phi(z_t,t,c)+(s-t)\widehat{\mathcal{D}}_t u_\theta(z_t,t,s,c)
 \right]\right\|_2^2
 \right],
 \label{eq:meanflow-loss}
\end{equation}
where $\operatorname{sg}$ denotes stop-gradient and $D$ counts scalar pixel values across all generated items in a group. We optimize only this average-velocity objective, without an auxiliary instantaneous-velocity loss. DDE takes differences between nearby predictions, making its accuracy sensitive to both arithmetic precision and perturbation size. We observe low-precision derivative errors in reference implementations and our own model, and therefore use higher precision for derivative estimation while retaining low-precision computation elsewhere in training to control the additional cost. \paperref{Appendix}{app:precision} examines the basis for this precision allocation.

At inference, the student predicts the average velocity for each interval and updates the items as
\begin{equation}
 z_{t_{k+1}}=z_{t_k}+(t_{k+1}-t_k)u_\theta(z_{t_k},t_k,t_{k+1},c),
 \label{eq:meanflow-sampling}
\end{equation}
starting from Gaussian noise at $t_0=0$ and reaching $t_4=1$ in four steps. \paperref{Table}{tab:main-results} shows that, on Text Refinement Bench, the four-step model reduces the average GPU generation cost per image to about one fifth of the 25-step teacher's cost, with a decrease of only 1.21 percentage points in Pass.

\section{Experiments}
\subsection{Experimental Setup}
AnyText Bench focuses on reconstructing annotated text at specified locations \citep{tuo2024anytext}, leaving a gap in open, reproducible evaluation of localized text refinement on generated images. We therefore construct \emph{Text Refinement Bench} from actual model-generated images, including those released in the public Qwen-Image Bench \citep{li2026qwen}. Human review verifies erroneous character regions and correction targets and excludes invalid cases before freezing the evaluation list, directly reflecting the practical need to correct local errors while preserving the original image. The benchmark evaluates correction accuracy and image preservation on 1,034 images with 2,638 repair targets. We also evaluate text reconstruction on the Chinese and English subsets of AnyText Bench to assess performance on conventional visual text generation. During evaluation, we identify inaccurate text polygons and failures of the original OCR evaluation pipeline on vertically arranged text. We correct the affected text regions and adapt OCR preprocessing for vertical layouts, reducing scoring errors caused by cropping and text orientation so that evaluation better reflects the quality of the generated text.

Standard OCR may transcribe a generated glyph with missing or extra strokes as the intended character, so correct recognition does not necessarily imply a successful repair. We therefore use CJK-CLIP, a dedicated CLIP-style glyph recognizer for characters,\footnote{The work presenting CJK-CLIP is currently under review; the model will be publicly released.} to distinguish correct glyphs from visually similar but structurally incorrect outputs for character-level evaluation on Text Refinement Bench. \paperref{Appendix}{app:benchmark} further describes the benchmark, its differences from related benchmarks, and the evaluation protocol.
\paragraph{Implementation details.}
We train the model on eight H800 GPUs using an internal Chinese--English dataset containing approximately 2.53 million image records. Base-model training uses AdamW with a learning rate of $10^{-4}$, $\beta=(0.9,0.95)$, zero weight decay, and an EMA decay of 0.9999. Each optimizer reinitialization starts with 2,000 linear warmup steps, followed by a constant learning rate. We follow a progressive-resolution schedule: we first train for approximately 13 epochs with at most $512^2$ pixels per main image, then increase this limit to $1024^2$ and continue training for approximately 15 epochs. We subsequently perform approximately 3,000 MeanFlow updates with an effective batch size of 24.

PixText uses patch size $p=16$, 32 Group JiT Blocks, hidden dimension 1280, and 16 attention heads. The denominator floor for velocity conversion is $\delta=0.05$. Initial logit-normal timestep sampling uses $\mu=-0.8$ and $\sigma=0.8$, followed by uniform sampling.

MeanFlow uses the same optimizer settings. Both the student and frozen teacher are initialized from the base model's EMA weights, and training intervals are formed by sorting two independently sampled uniform timesteps. The default DDE step size is $h=0.005$, reduced near boundaries as described above. Finite-difference predictions and subtraction use FP32 tensor arithmetic, with TF32 enabled for matrix multiplications; the trainable student forward and backward passes use BF16 mixed precision. For the main-table evaluation, PixText uses 25 Euler steps and MeanFlow uses four Euler steps, both with a glyph-conditioned classifier-free guidance (CFG) \citep{ho2022classifier} scale of 3.3 (see \paperref{Appendix}{app:glyph-cfg} for details).

\subsection{Quantitative Comparisons}
\paperref{Table}{tab:main-results} compares eight text-specialized methods with PixText and its four-step MeanFlow variant. On Text Refinement Bench, PixText achieves a Pass of 83.06\%, compared with 36.77\% for the strongest baseline, FLUX-Text, while also attaining the lowest FID \citep{heusel2017gans} and LPIPS \citep{zhang2018unreasonable}. PP-OCRv6 gives PixText 94.66\% exact target-character accuracy, compared with 52.31\% for FLUX-Text.

On AnyText Bench, PixText achieves the highest Chinese text accuracy and ranks second to FLUX-Text on English, showing that its refinement advantage is accompanied by competitive text reconstruction. For each Chinese-capable baseline, we report the complete line- or character-input run with higher text accuracy; \paperref{Table}{tab:anytext-zh-input} reports both settings and all four metrics. The four-step MeanFlow variant retains the refinement advantage over the baselines while reducing generation cost relative to the base PixText model, making localized correction practical in generation workflows. \paperref{Appendix}{app:qwen-api-refinement} reports a comparison with Qwen-Image-3.0-Pro through its editing API on Text Refinement Bench.

\begin{figure}[!t]
 \centering
 {\setlength{\tabcolsep}{0pt}\fontsize{7.5}{8.5}\selectfont
 \newcommand{\qualheadcell}[1]{\makebox[0.099286\linewidth][c]{#1}}
 \hspace*{0.003571\linewidth}%
 \begin{tabular}{@{}*{10}{c}@{}}
 \qualheadcell{Original} & \qualheadcell{Input} & \qualheadcell{Target} &
 \qualheadcell{TextDiffuser} & \qualheadcell{AnyText} & \qualheadcell{AnyText2} &
 \qualheadcell{FLUX-Text} & \qualheadcell{TextFlux} &
 \cellcolor[HTML]{DDF3FC}\qualheadcell{\bfseries PixText} &
 \cellcolor[HTML]{DDF3FC}\qualheadcell{\bfseries PixText +}\\
 & & & & & & & & \cellcolor[HTML]{DDF3FC} &
 \cellcolor[HTML]{DDF3FC}\qualheadcell{\bfseries MeanFlow}
 \end{tabular}\par}\vspace{1pt}
 \includegraphics[trim=0 0 0 28.8bp,clip,width=\linewidth]{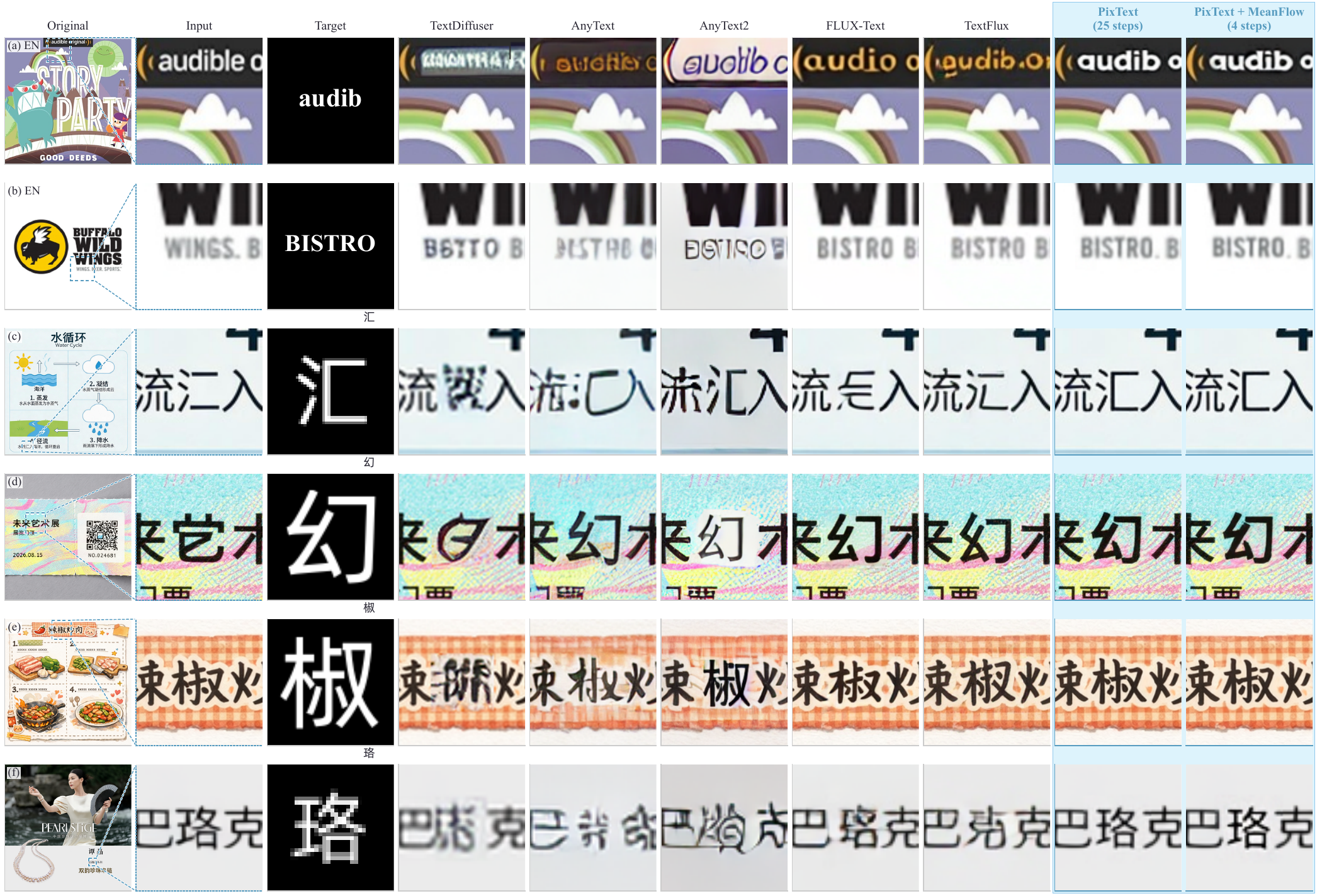}
 \caption{Qualitative comparisons of different methods.}
 \label{fig:qualitative-main}
\end{figure}
\begin{table}[t]
\centering
\caption{Text Refinement Bench and AnyText results. AnyText ZH uses each baseline's higher-Acc run for all metrics; Appendix D compares inputs. Dashes mean unsupported. Params (B) include the generator, VAE, and text encoders. Bold and underline mark best and second best.}
\label{tab:main-results}
\begingroup
\definecolor{pixtextsky}{HTML}{DDF3FC}
\scriptsize
\setlength{\tabcolsep}{2pt}
\renewcommand{\arraystretch}{1.15}
\resizebox{\linewidth}{!}{%
\begin{tabular}{l||rrrrr|rrrr|rrrr|rr}
\hline\hline
\rowcolor{gray!20}
 & \multicolumn{5}{c|}{\textbf{Text Refinement Bench}} & \multicolumn{4}{c|}{\textbf{AnyText EN}} & \multicolumn{4}{c|}{\textbf{AnyText ZH}} & \multicolumn{2}{c}{\textbf{Efficiency}} \\
\rowcolor{gray!20}
Method & Pass$\uparrow$ & R@1$\uparrow$ & \shortstack{OCR Acc$\uparrow$} & FID$\downarrow$ & LPIPS$\downarrow$ & Acc$\uparrow$ & NED$\uparrow$ & FID$\downarrow$ & LPIPS$\downarrow$ & Acc$\uparrow$ & NED$\uparrow$ & FID$\downarrow$ & LPIPS$\downarrow$ & \shortstack{GPU-s/img$\downarrow$} & Params (B)$\downarrow$ \\
\hline\hline
TextDiffuser & -- & -- & -- & -- & -- & 54.15 & 0.7863 & 24.04 & 0.1443 & -- & -- & -- & -- & \underline{3.126} & 1.066 \\
\rowcolor{gray!10}
TextCtrl & -- & -- & -- & -- & -- & 59.53 & 0.8378 & 8.16 & 0.0365 & -- & -- & -- & -- & 13.802 & 1.128 \\
UDiffText & -- & -- & -- & -- & -- & 67.16 & 0.8599 & 16.94 & 0.1072 & -- & -- & -- & -- & 8.252 & 1.361 \\
\rowcolor{gray!10}
DiffSTE & -- & -- & -- & -- & -- & 61.01 & 0.8019 & 19.16 & 0.1258 & -- & -- & -- & -- & 16.132 & 1.315 \\
AnyText & 12.40 & 23.46 & 25.59 & 11.47 & 0.0187 & 64.60 & 0.8440 & 22.40 & 0.1111 & 66.54 & 0.8607 & 22.29 & 0.0949 & 4.310 & 1.428 \\
\rowcolor{gray!10}
AnyText2 & 27.29 & 43.56 & 48.41 & 10.50 & 0.0256 & 79.65 & 0.9132 & 32.15 & 0.1752 & 78.06 & 0.9112 & 28.89 & 0.1407 & 8.871 & 1.485 \\
FLUX-Text & 36.77 & 48.26 & 52.31 & 2.59 & 0.0028 & \textbf{84.05} & \textbf{0.9365} & 11.50 & 0.0684 & 80.93 & 0.9233 & 12.22 & 0.0498 & 16.420 & 17.799 \\
\rowcolor{gray!10}
TextFlux & 14.18 & 23.77 & 28.32 & 2.27 & 0.0023 & 74.50 & 0.8752 & 14.06 & 0.0856 & 65.06 & 0.8551 & 12.82 & 0.0576 & 14.144 & 16.872 \\
\hline
\rowcolor{pixtextsky}
\textbf{PixText} & \textbf{83.06} & \textbf{92.95} & \textbf{94.66} & \textbf{0.34} & \textbf{0.0010} & \underline{80.91} & \underline{0.9190} & \textbf{7.62} & \underline{0.0295} & \textbf{84.13} & \textbf{0.9477} & \underline{5.93} & \underline{0.0199} & 4.514 & \textbf{0.952} \\
\rowcolor{pixtextsky}
\textbf{PixText + MeanFlow} & \underline{81.84} & \underline{92.57} & \underline{94.58} & \underline{0.38} & \underline{0.0011} & 80.75 & 0.9142 & \underline{7.80} & \textbf{0.0293} & \underline{82.74} & \underline{0.9429} & \textbf{5.92} & \textbf{0.0193} & \textbf{0.942} & \underline{0.954} \\
\hline\hline
\end{tabular}}
\endgroup
\end{table}

\subsection{Qualitative Comparisons}
\paperref{Figure}{fig:qualitative-main} compares six localized text refinement or generation cases across lettering styles and backgrounds. At matched scales, PixText and its four-step MeanFlow variant produce legible text while preserving local appearance; some baselines distort strokes, style, or nearby content. \paperref{Appendix}{app:qualitative} gives fourteen more cases, and \paperref{Appendix}{app:ood-generation} shows glyph-conditioned generation beyond the main Chinese--English training scripts.

\section{Conclusion}
PixText refines text through localized pixel-space grouped generation, preserving fine strokes and surrounding content. It substantially improves correction on Text Refinement Bench, while its four-step MeanFlow variant costs less than one amortized GPU-second per image. Targeted pixel-space generation thus improves text readability while preserving image appearance.

\subsection*{AI use statement}
We used generative AI tools to assist with manuscript polishing and translation, coding assistance, and experimental data organization. The authors take responsibility for the final paper and reported results.

\bibliographystyle{iclr2027_conference}
\bibliography{references}

\clearpage
\appendix
\section{Variable-Length Batch Training with FlexAttention}
\label{app:flex}
The main-image resolution, number of edit regions, and region sizes vary across samples, giving each image group a different number of tokens. We concatenate the main-image, item-image, and glyph tokens from all groups along the sequence dimension to form a batch, without padding each group to a common length. Each token retains its group membership and spatial position within the group, allowing samples with different image sizes and numbers of regions to be trained together.

The attention mask controls both isolation between groups and interaction within each group. Attention between different groups is fully masked. Within a group, the main image and items interact bidirectionally, items attend to one another, and each glyph interacts across images only with its paired item; self-attention within each image is retained. Sequence concatenation therefore changes how the batch is organized while preserving the information flow within each sample.

We encode these rules as a FlexAttention block-sparse mask and reuse it across Transformer layers. FlexAttention skips fully masked attention blocks and applies the specified masking within the remaining blocks, without explicitly storing a dense attention matrix for the entire concatenated sequence. Combining variable-length sequence packing with masked attention avoids the computation introduced by padding to the longest sample, while preserving joint modeling within groups and independence between groups.

\section{Glyph-Conditioned Classifier-Free Guidance}
\label{app:glyph-cfg}
\paragraph{Condition dropout.}
We train the two classifier-free guidance (CFG) branches by randomly dropping glyph conditions. For each training image group, we remove all glyph tokens associated with its items with probability 0.1; otherwise, all glyph conditions are retained. Dropping a condition removes the glyph branch from the input sequence. The main image, noisy items, and region geometry remain available, so ``unconditional'' here refers only to the absence of target glyphs; scene context is still provided. Random condition dropout is disabled during validation and inference, when predictions with and without glyph conditions are constructed explicitly.

\paragraph{Inference guidance.}
We separate the retained scene and geometry conditions $c_{\mathrm{ctx}}$ from the glyph conditions $g$. Let $\hat{x}_{1}^{\mathrm{cond}}=f_\theta(z_t,t,c_{\mathrm{ctx}},g)$ and $\hat{x}_{1}^{\mathrm{uncond}}=f_\theta(z_t,t,c_{\mathrm{ctx}},\emptyset)$. The guided clean-item prediction is
\begin{equation}
 \hat{x}_{1}^{\mathrm{CFG}}=
 \hat{x}_{1}^{\mathrm{uncond}}+
 \gamma(t)\bigl(\hat{x}_{1}^{\mathrm{cond}}-\hat{x}_{1}^{\mathrm{uncond}}\bigr).
 \label{eq:glyph-cfg}
\end{equation}
Both predictions share the noisy state, timestep, main image, and region positions. By default, the sampler uses the specified guidance scale for $0.1<t<1$ and sets $\gamma(t)=1$ otherwise, returning the conditional prediction directly. The main-table guidance scale is 3.3.

\paragraph{Use in MeanFlow.}
MeanFlow training retains the same group-level glyph dropout, and the teacher and student receive the same glyph-condition state. The teacher predicts the target instantaneous velocity under the sampled glyph-condition state. At inference, both branches share the interval start $t$ and endpoint $s$. Their interval-conditioned clean predictions are combined as above and then converted to average velocity. Because both branches share the noisy state and velocity-conversion denominator, this is equivalent to
\begin{equation}
 u_{\theta}^{\mathrm{CFG}}=
 u_{\theta}^{\mathrm{uncond}}+
 \gamma(t)\bigl(u_{\theta}^{\mathrm{cond}}-u_{\theta}^{\mathrm{uncond}}\bigr),
 \label{eq:meanflow-glyph-cfg}
\end{equation}
followed by an Euler update. Thus, four steps denotes four interval updates; steps with CFG enabled require separate predictions with and without glyph conditions.

\section{Precision and Efficiency of MeanFlow DDE}
\label{app:precision}
DDE estimates derivatives by subtracting nearby predictions, making the estimate sensitive to the numerical precision of endpoint evaluation. We use FP32 tensors for endpoint predictions and subtraction, enable TF32 matrix multiplications, and retain BF16 computation for the trainable center forward and backward passes.

We compare BF16, strict FP32, and our mixed-precision endpoint computation on the MeanFlow model used in the main results, using three training samples with a fixed interval $(t,s)=(0.4,0.8)$ and finite-difference step size $h=0.005$. The reference $D_{\mathrm{FP32}}$ is the finite-difference estimate computed in strict FP32 with TF32 disabled; relative error is $\|D-D_{\mathrm{FP32}}\|_2/\|D_{\mathrm{FP32}}\|_2$. Timing on dedicated NVIDIA H800 GPUs covers derivative estimation, the center forward pass, loss computation, and backward propagation, with two warm-up runs followed by seven measurements per configuration.

\begin{table}[t]
 \centering
 \caption{DDE precision and efficiency. All configurations use BF16 center forward and backward passes and differ only in endpoint computation. We report medians over three samples, taking the median of seven timing measurements for each sample.}
 \label{tab:dde-precision}
 \begin{tabular}{lrr}
  \toprule
  Endpoint computation & Relative error (\%) $\downarrow$ & Time (s) $\downarrow$ \\
  \midrule
  BF16 & 192.64 & 0.551 \\
  Strict FP32 (reference) & 0.00 & 23.895 \\
  Ours (mixed precision) & \textbf{18.92} & \textbf{0.706} \\
  \bottomrule
 \end{tabular}
\end{table}

Table~\ref{tab:dde-precision} shows that our mixed-precision scheme reduces the median relative derivative error from 192.64\% to 18.92\%, with only 28\% more computation time than BF16, while running approximately $34\times$ faster than strict FP32. This supports concentrating higher precision on derivative estimation to reduce finite-difference error while keeping computation efficient.

\section{Text Refinement Bench and Evaluation Settings}
\label{app:benchmark}
\paragraph{Comparison with existing benchmarks.}
Text Refinement Bench targets local character errors in generated images, complementing text-generation benchmarks such as AnyText \citep{tuo2024anytext} and instruction-driven editing benchmarks such as TextEdit, TextEditBench, and MultiTextEdit \citep{tian2026internvl,gui2025texteditbench,cheng2026multitextedit}. Explicit repair regions and target characters focus evaluation on correction quality and image preservation, reducing dependence on instruction interpretation and target localization. ScenePair \citep{zeng2024textctrl} also provides region-level pairs, but its source text and target reference come from the same image. Although its original crop-based protocol does not expose the reference, directly supplying the full image as context would reveal it; evaluating context-aware refinement therefore requires adapting this setup. Our benchmark is built around repairing actual generation errors with the surrounding image as context, and jointly measures target-glyph agreement, character identification, and image preservation.
\paragraph{Evaluation set.}
The reviewed benchmark contains 1,034 images and 2,638 repair targets. All methods use all 2,638 repair targets for character-level metrics, and FID and LPIPS cover the same 1,034 images.
\paragraph{Evaluating glyph correctness.}
We use CJK-CLIP to measure similarity between a character image and its intended character, evaluating glyph correctness beyond whether the image can be recognized as the target.
\paragraph{Pass and R@1.}
All methods are evaluated with the same CJK-CLIP model and character-specific pass thresholds. Let $S(I,y)$ denote the similarity between character image $I$ and target $y$, and let $\mu_y$ be its mean over reference images of that character. A repaired glyph passes when
\begin{equation}
 \operatorname{pass}(I,y)=\mathbb{1}\!\left[S(I,y)\ge\mu_y-0.03\right].
 \label{eq:glyph-pass}
\end{equation}
Pass is the fraction of target characters satisfying this criterion; R@1 is the fraction for which the target ranks first by similarity over the character vocabulary. They measure target-glyph agreement and character identification, respectively. A malformed glyph can still be closest to the intended character, explaining why R@1 can exceed Pass. AnyText Bench retains its OCR-based evaluation.
\paragraph{OCR check.}
We additionally evaluate the same 2,638 target characters with the PP-OCRv6 medium recognizer.\footnote{\url{https://huggingface.co/PaddlePaddle/PP-OCRv6_medium_rec}} OCR Acc is the percentage whose transcription exactly matches the repair target; blank or multi-character outputs count as errors. OCR Acc measures recognizability, whereas Pass additionally tests glyph fidelity.

\paperref{Figure}{fig:ocr-glyph-cases} illustrates stroke errors missed by high-confidence OCR recognition. All four glyphs are transcribed as their targets with confidence between 99.67\% and 99.97\%, yet fail the CJK-CLIP criterion, consistent with the human error annotations.
\begin{figure}[t]
  \centering
  \includegraphics[width=\linewidth]{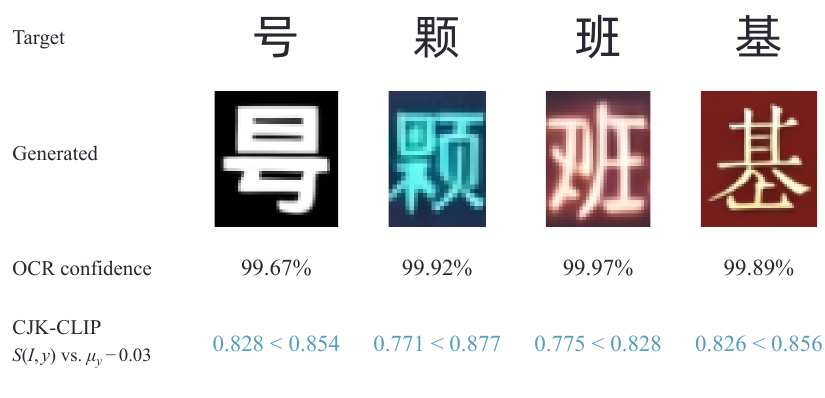}
  \caption{Correct recognition does not imply correct glyphs. OCR confidently transcribes these malformed glyphs as the targets above, CJK-CLIP similarity scores fall below their respective acceptance thresholds $\mu_y-0.03$, yielding Error decisions consistent with human annotations.}
  \label{fig:ocr-glyph-cases}
\end{figure}

\paragraph{Metrics.}
Pass, R@1, and OCR Acc on Text Refinement Bench, and sentence accuracy (Acc) on AnyText, are reported as percentages. LPIPS uses AlexNet on full images resized to $256\times256$. Chinese AnyText OCR scores use a common set of vertical text boxes and symbol-width equivalence rules.
\paragraph{AnyText evaluation corrections.}
Some original polygons do not accurately cover the text, so direct cropping can truncate glyphs or include neighboring content. We review the character boxes and use the corrected regions for vertically arranged Chinese text. Characters are cropped in top-to-bottom reading order, kept upright, and concatenated horizontally before OCR, avoiding recognition errors caused by rotating the entire vertical line. All compared Chinese methods use the same reviewed annotations and recognition preprocessing; horizontal text retains the original line-level crops.
\paragraph{Input settings.}
For AnyText ZH, we evaluate the four Chinese-capable baselines with both the official line polygons and the character regions supplied to PixText. For each method, the main table reports the complete run with higher sentence accuracy: character input for AnyText and line input for AnyText2, FLUX-Text, and TextFlux. NED, FID, and LPIPS come from that same run. TextDiffuser, TextCtrl, UDiffText and DiffSTE do not support Chinese target conditioning in the evaluated official checkpoints. We therefore report their AnyText English results on all 1,000 images and 4,181 lines, and mark AnyText ZH and the Chinese-containing Text Refinement Bench as unsupported. UDiffText uses crop-based editing with polygon-restricted compositing; DiffSTE uses a 256-pixel working canvas and sequential target editing. Dashes indicate unsupported settings.

\begin{table*}[t]
\centering
\caption{AnyText ZH with official line inputs and PixText's character-level inputs. All rows use the same 1,000-image benchmark and 2,092-line scoring protocol. The input with higher sentence accuracy (Acc) is marked $\star$ and supplies all four metrics in the main table; metrics are never selected independently.}
\label{tab:anytext-zh-input}
\small
\setlength{\tabcolsep}{9pt}
\begin{tabular}{llrrrr}
\toprule
Method & Input & Acc (\%)$\uparrow$ & NED$\uparrow$ & FID$\downarrow$ & LPIPS$\downarrow$ \\
\midrule
AnyText & Line & 65.87 & 0.8629 & 21.71 & 0.0891 \\
 & Character$\star$ & 66.54 & 0.8607 & 22.29 & 0.0949 \\
\addlinespace[3pt]
AnyText2 & Line$\star$ & 78.06 & 0.9112 & 28.89 & 0.1407 \\
 & Character & 71.94 & 0.8948 & 28.52 & 0.1353 \\
\addlinespace[3pt]
FLUX-Text & Line$\star$ & 80.93 & 0.9233 & 12.22 & 0.0498 \\
 & Character & 78.49 & 0.9191 & 12.34 & 0.0562 \\
\addlinespace[3pt]
TextFlux & Line$\star$ & 65.06 & 0.8551 & 12.82 & 0.0576 \\
 & Character & 63.81 & 0.8493 & 13.11 & 0.0614 \\
\bottomrule
\end{tabular}
\end{table*}

\paperref{Table}{tab:anytext-zh-input} shows that all four Chinese-capable baselines can use character-level targets. AnyText improves from 65.87\% to 66.54\% Acc with character input; the other three baselines achieve higher Acc with their line inputs. Reporting the stronger setting for each method preserves its complete text and image metrics while making the input choice explicit.
\paragraph{Computational cost.}
GPU-s/img is the summed GPU generation time divided by the number of images. Batch sizes are chosen to keep GPU compute units fully occupied throughout generation. Parameter counts include the generation network, VAE, and text encoders. Generation networks include control branches and unmerged adapters.

On one NVIDIA H800, we time batch-one requests for 32 evenly spaced images from Text Refinement Bench, with two warm-up runs per implementation and three measurements per image. PixText + MeanFlow averages 1.152 s per request; batching the conditional and unconditional CFG branches reduces this to 1.062 s (7.8\% faster). Even with one image per request, PixText + MeanFlow remains faster than all compared baselines, whose lowest amortized cost is 3.126 GPU-s/img.
\section{Compact Crops and Token Savings Enabled by CCPE}
\label{app:ccpe-tokens}
CCPE allows item crops to start independently of the main-image patch grid; only their width and height must be multiples of the patch size. Compared with expanding each text box outward to grid boundaries, this reduces the extra border area included for alignment. \paperref{Figure}{fig:ccpe-tokens} illustrates both crops for the same text region.

We compare the two crop choices on 116,038 text regions from 5,000 images randomly sampled from our training dataset, using a patch size of 16 pixels. Grid alignment requires 2,923,250 item tokens, whereas CCPE requires 2,523,979, a reduction of \textbf{13.7\%}. Both cover the same target text; the savings come from tighter crops.

\begin{figure}[!ht]
  \centering
  \includegraphics[width=\linewidth]{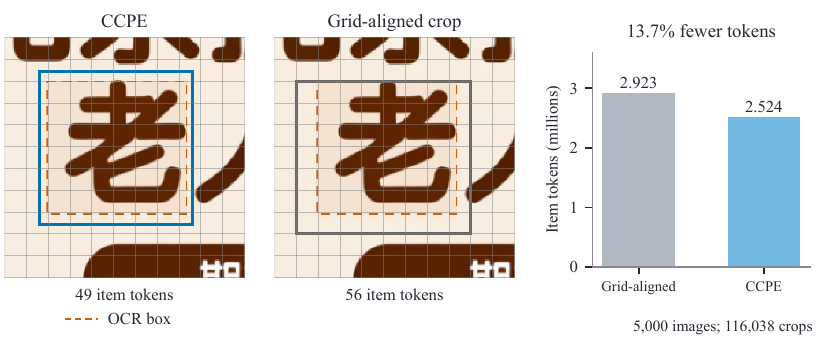}
  \caption{CCPE reduces crop padding. Left and middle: the same text region with two crop boundaries; orange dashed boxes mark the original OCR regions, and the background grid has 16-pixel spacing. Right: total item tokens across 116,038 text regions.}
  \label{fig:ccpe-tokens}
\end{figure}

\section{Data Preparation and Sampling Details}
\label{app:training-sampling}
\paragraph{Data preparation.} We use CRAFT to obtain text regions and PaddleOCR to recognize their contents for constructing the glyph conditions used in training \citep{baek2019character,cui2025paddleocr}.
\paragraph{Timestep scheduling.}
This ablation uses an earlier model. It is trained for 340k steps at 512 resolution and then 120k steps at 1024. Both branches start from the same initial checkpoint using its EMA weights and sample either $t=\operatorname{sigmoid}(-0.8+0.8\epsilon)$, $\epsilon\sim\mathcal{N}(0,1)$, or $t\sim\mathcal{U}(0,1)$, with timestep shifting disabled. Both use an effective global batch of 32 and are evaluated after 5k and 10k continuation updates. The learning rate is $10^{-4}$ and EMA decay is 0.9999; the optimizer, 2k-step warmup, and data-sampling progress restart. The plotted horizontal axis counts continuation updates only.

\paragraph{MiniT2I comparison.}
We compare the official MiniT2I-B/16 logit-normal-fine-tuned checkpoint with the EMA from 40k uniform fine-tuning updates starting from released pretrained weights. The official fine-tuning configuration also specifies 40k updates. Evaluation uses 50 paired prompts and noise seeds at $512\times512$, 100 Euler steps, and CFG 2.5, with five inference schedules matched between models. Inference timesteps are constructed as $t=u/[\alpha+(1-\alpha)u]$, where $u$ is a uniform grid on $[0,1]$. Setting $\alpha=1$ leaves the grid unchanged; $\alpha>1$ shifts interior timesteps toward the noise endpoint ($t=0$), and $\alpha<1$ toward the data endpoint ($t=1$). We use the same normalized boundary-gradient metric and 20,000 paired prompt-bootstrap resamples. Without inference shifting, the score decreases from 1.2419 to 1.1511; the paired difference is $-0.0908$ with a 95\% interval of $[-0.1194,-0.0619]$. All five schedules have paired intervals below zero.

\paragraph{Paired statistics.}
For PixText, the uniform-minus-logit-normal difference is $-0.0573$ at 5k updates, with a 95\% interval of $[-0.0733,-0.0394]$, and $-0.0445$ at 10k, with an interval of $[-0.0703,-0.0184]$. Confidence intervals are computed by paired bootstrap over evaluation images.

\begin{figure}[p]
  \centering
  \includegraphics[width=\linewidth]{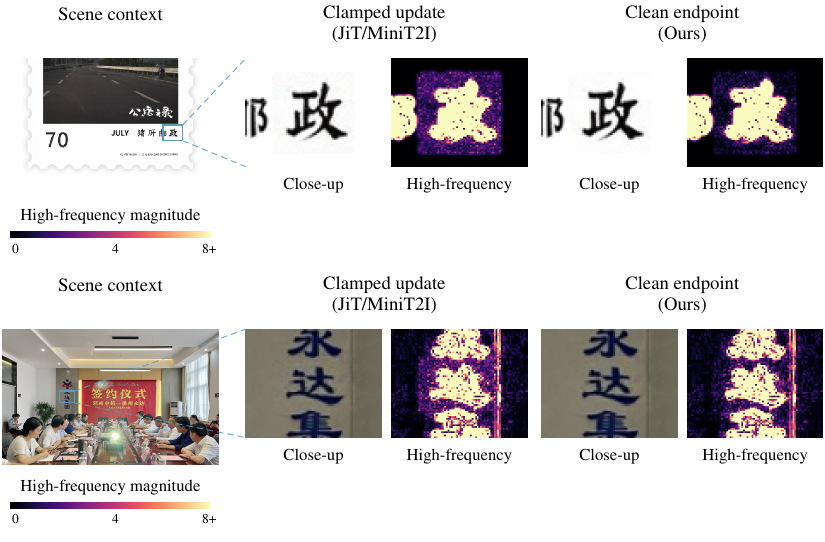}
  \caption{Additional endpoint-update comparisons. Each row shows scene context on the left and a close-up with its high-frequency map for each update on the right. Both maps share the color scale below the scene.}
  \label{fig:sampling-endpoint-more}
  \vspace{12pt}
  \includegraphics[width=\linewidth]{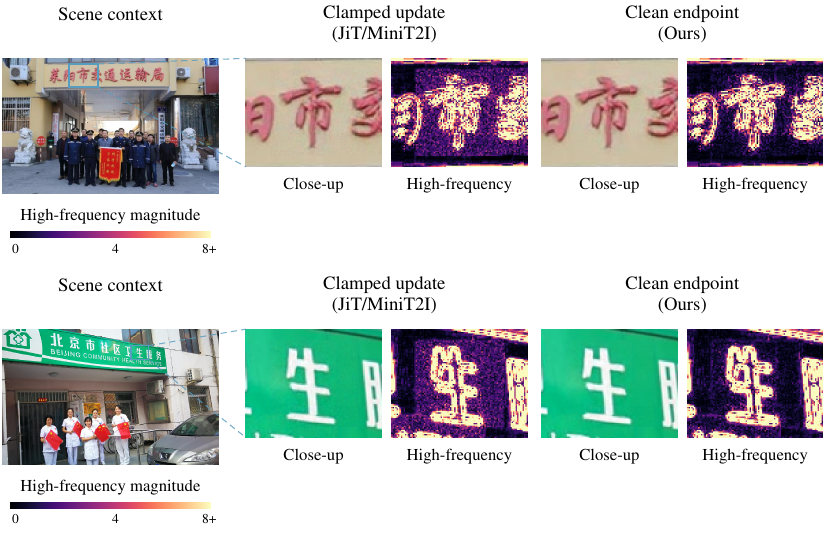}
  \caption{Additional endpoint-update comparisons (continued), with the same layout as \paperref{Figure}{fig:sampling-endpoint-more}.}
  \label{fig:sampling-endpoint-more-second}
\end{figure}

\paragraph{Grid-artifact measurement.}
Following the adjacent RGB-gradient probe described in the miniT2I blog,\footnote{\url{https://peppaking8.github.io/posts/minit2i.html}} we use the following explicit normalization. Let $\mathcal{E}$ contain all horizontal and vertical adjacent pixel pairs in a crop, let $\mathcal{E}_{16}$ contain those crossing a 16-pixel patch boundary, and let $g_e$ be the mean absolute RGB difference of pair $e$. We define
\begin{equation}
 S(I)=\frac{\operatorname{mean}_{e\in\mathcal{E}_{16}}g_e}
 {\operatorname{mean}_{e\in\mathcal{E}}g_e}.
 \label{eq:grid-gradient}
\end{equation}
We evaluate 128 images from the fixed Text Refinement Bench list, with fixed target text and noise seeds. All checkpoints use EMA weights, BF16, 25 unshifted Euler steps, and CFG 2.2 active on $[0.1,1.0]$. We measure 8-bit RGB crops at their original resolution before resizing or compositing. Applying the minimum height and width of 32 pixels leaves 297 crops from 126 images. Scores are averaged over crops within each image, then equally over images; confidence intervals use 4,000 paired image-bootstrap resamples. Source crops are scored identically as a content reference.
\paragraph{Sampling endpoint.}
We fix the model weights, validation samples, and noise seeds, and compare two final updates on identical 50-step Heun trajectories. Sampling uses BF16, CFG 2.2 active on $[0.1,1.0]$, and no timestep shifting. The baseline converts predictions to velocity using the denominator $\max(1-t,0.05)$; the alternative returns the guided clean prediction only at the final step. For the final transition from $t=0.98$ to 1, the step size is 0.02, while the velocity denominator remains clamped to 0.05. The baseline therefore moves only $0.02/0.05=40\%$ of the way from $x_t$ toward the clean prediction, yielding $x_{\mathrm{next}}=0.6x_t+0.4\hat{x}_1$. The final output thus retains a 60\% contribution from the current noisy state instead of fully reaching the clean prediction, explaining the residual speckle. Returning $\hat{x}_1$ directly completes this final denoising update. \paperref{Figures}{fig:sampling-endpoint-more} and~\ref{fig:sampling-endpoint-more-second} supply four additional cases.

\paragraph{Speckle visualization.}
We compute the mean absolute RGB difference between each full output and its Gaussian-smoothed version, then extract the same region for visualization. The Gaussian standard deviation is one native pixel; both maps share a linear range of 0--8 in 8-bit RGB units, with larger values saturated. The maps visualize local high-frequency content, including text edges and fine background fluctuations. The displayed output images are unfiltered.

\clearpage
\section{Additional Qualitative Comparisons}
\label{app:qualitative}
\begin{figure}[!ht]
 \centering
 \includegraphics[width=0.96\linewidth]{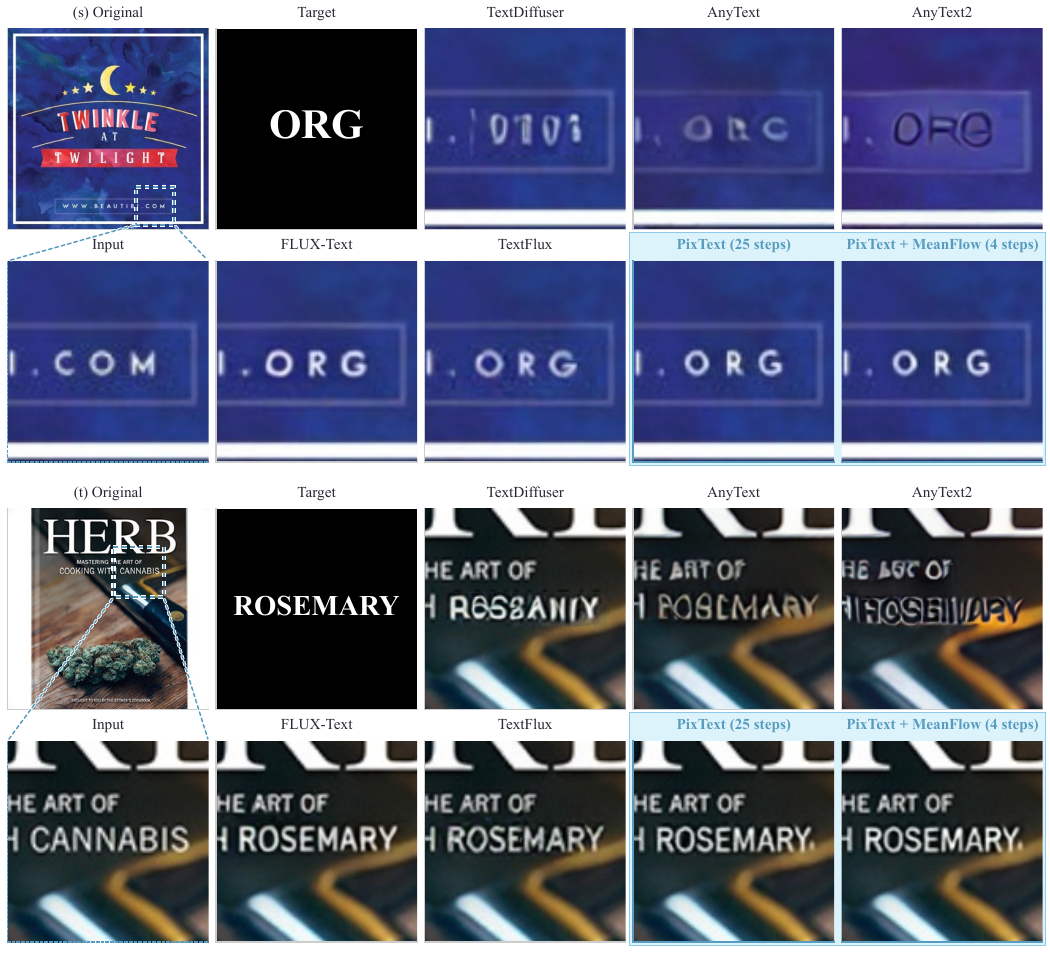}
 \caption{Additional qualitative comparisons of different methods (5/5), using the same panel order, crop coordinates, and display scales as \paperref{Figure}{fig:qualitative-extra-1}.}
 \label{fig:qualitative-extra-5}
\end{figure}

\paperref{Figures}{fig:qualitative-extra-1}--\ref{fig:qualitative-extra-5} present fourteen additional qualitative comparisons involving small text, varied stroke styles, perspective, and textured backgrounds. The figures compare the same seven methods as \paperref{Figure}{fig:qualitative-main}. Each two-row group shows an annotated target region, the requested text, and the corresponding input and method outputs, with identical crop coordinates and display scales across methods.
\begin{figure}[!ht]
 \centering
 \includegraphics[width=0.96\linewidth]{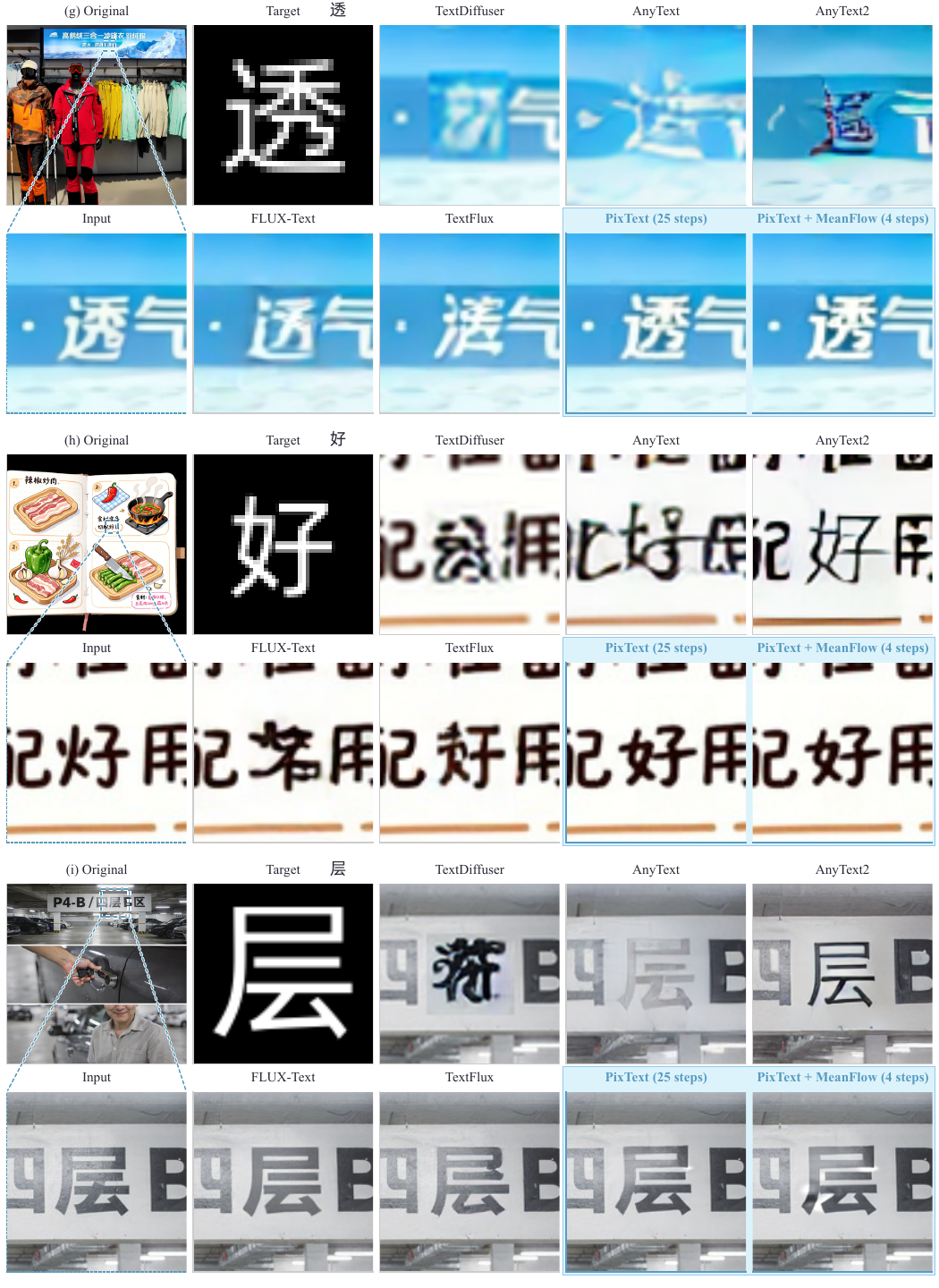}
 \caption{Additional qualitative comparisons of different methods (1/5).}
 \label{fig:qualitative-extra-1}
\end{figure}
\clearpage
\begin{figure}[!ht]
 \centering
 \includegraphics[width=0.96\linewidth]{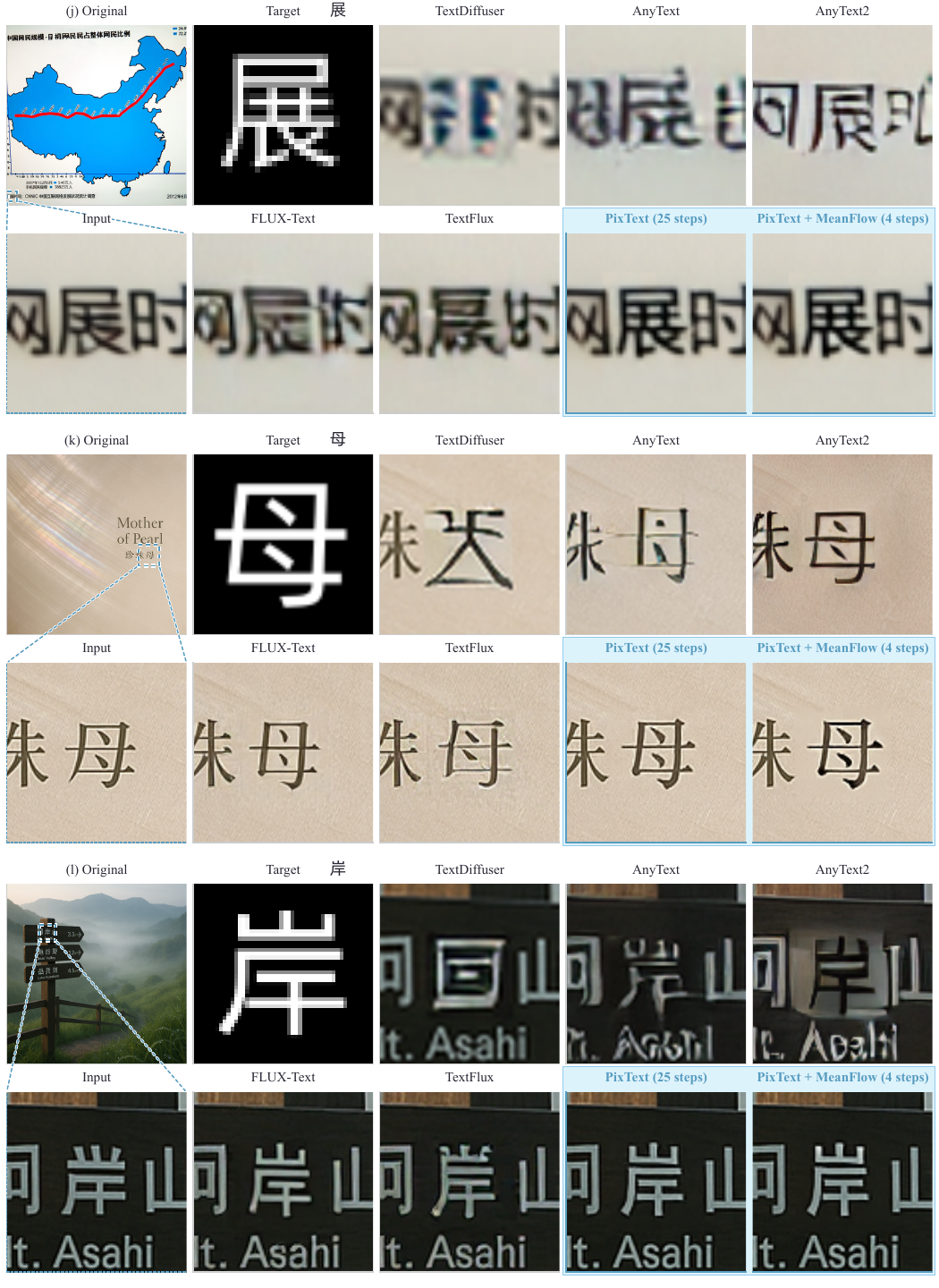}
 \caption{Additional qualitative comparisons of different methods (2/5), using the same panel order, crop coordinates, and display scales as \paperref{Figure}{fig:qualitative-extra-1}.}
 \label{fig:qualitative-extra-2}
\end{figure}
\clearpage
\begin{figure}[!ht]
 \centering
 \includegraphics[width=0.96\linewidth]{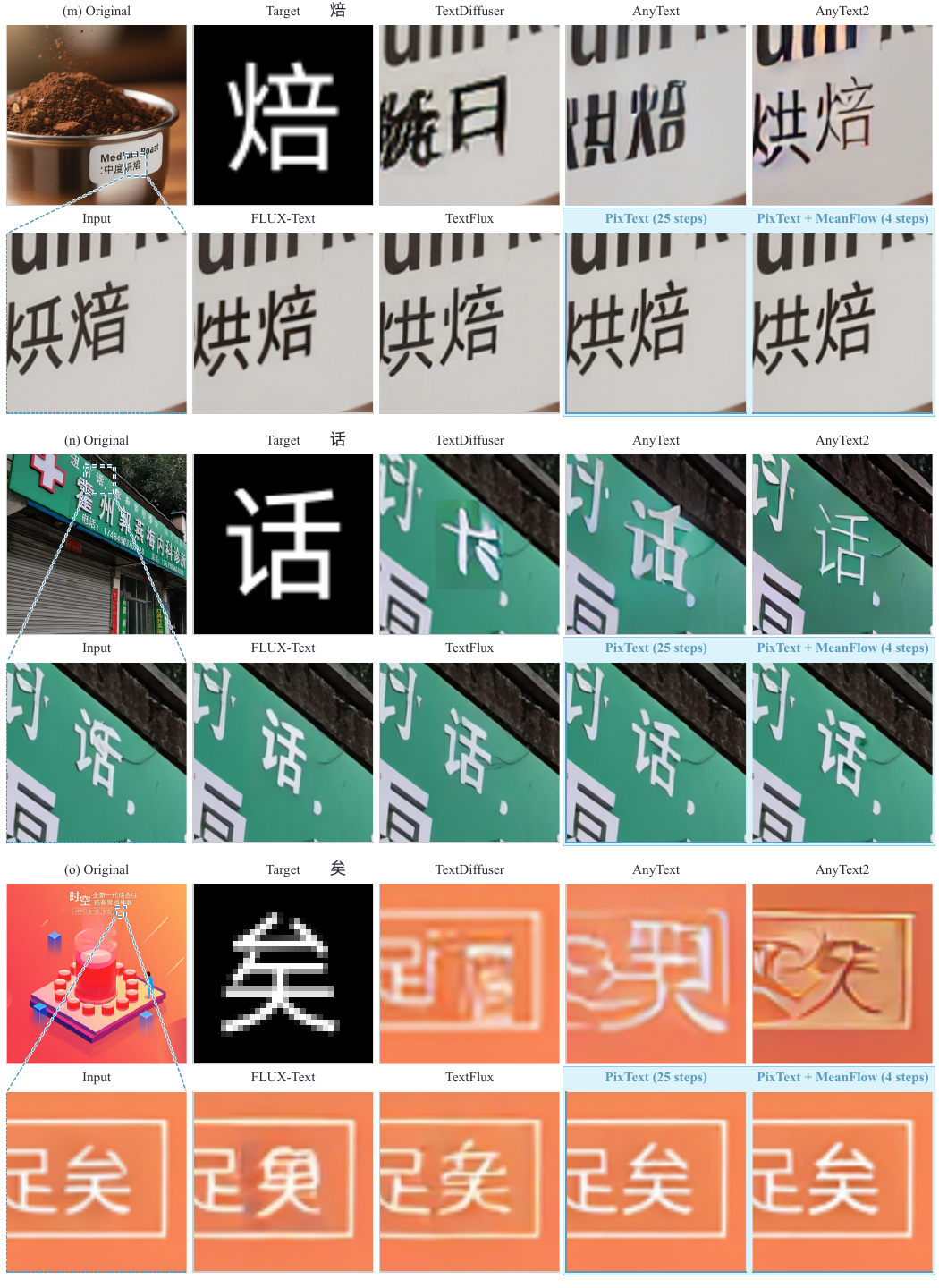}
 \caption{Additional qualitative comparisons of different methods (3/5), using the same panel order, crop coordinates, and display scales as \paperref{Figure}{fig:qualitative-extra-1}.}
 \label{fig:qualitative-extra-3}
\end{figure}
\clearpage
\begin{figure}[!ht]
 \centering
 \includegraphics[width=0.96\linewidth]{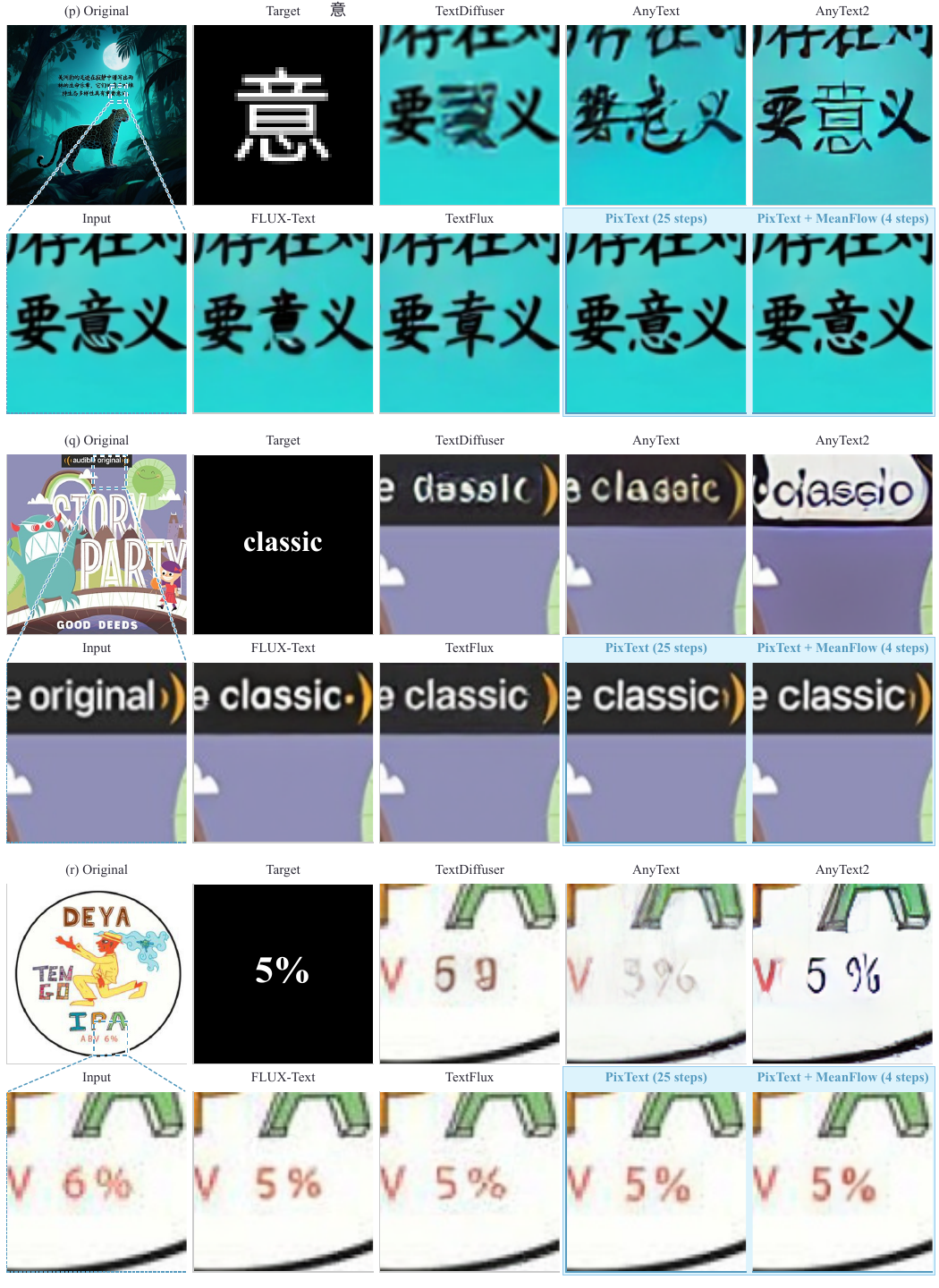}
 \caption{Additional qualitative comparisons of different methods (4/5), using the same panel order, crop coordinates, and display scales as \paperref{Figure}{fig:qualitative-extra-1}.}
 \label{fig:qualitative-extra-4}
\end{figure}
\clearpage

\clearpage
\section{Generation beyond the Main Training Scripts}
\label{app:ood-generation}
Although PixText is trained primarily on Chinese and English data, glyph conditions directly describe the visual shapes of target characters, providing a natural representation for specifying other characters. In the examples shown in \paperref{Figure}{fig:ood-generation}, the model follows the supplied glyphs to generate the target content while retaining adjacent text and background. These qualitative results suggest the potential of glyph conditioning for generation beyond the main training scripts.
\begin{figure}[!ht]
  \centering
  \includegraphics[width=\linewidth]{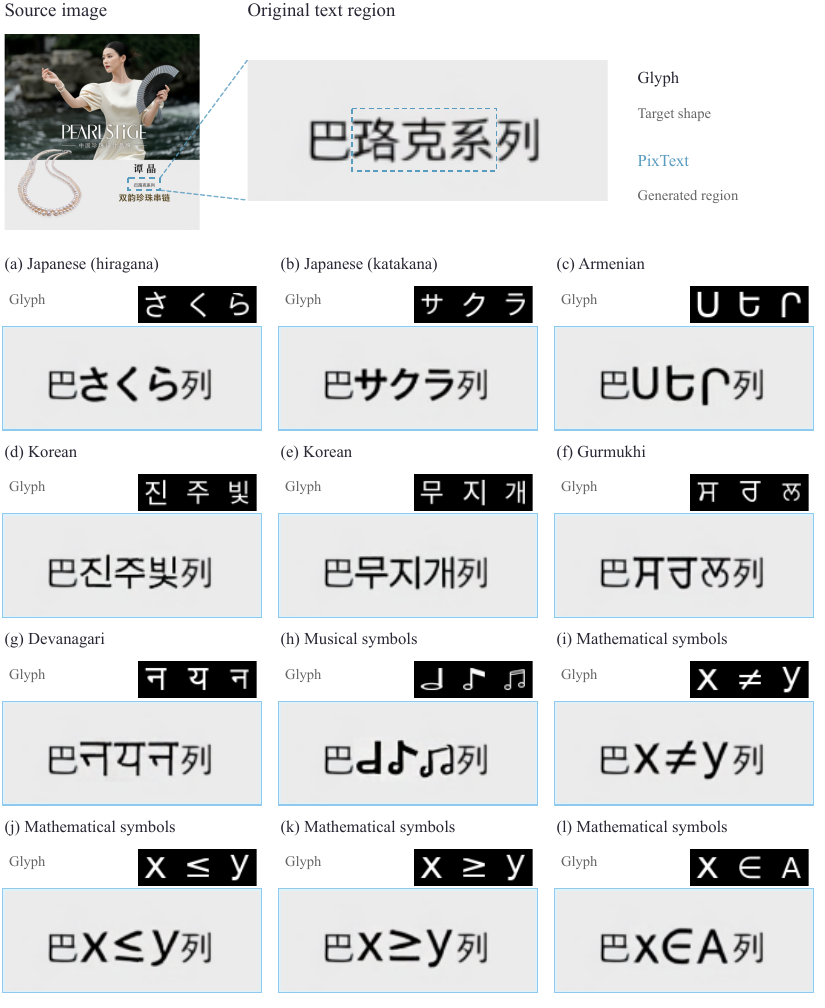}
  \caption{Generation beyond the main training scripts. Top: shared source image and target region. (a--l) Glyph conditions and the corresponding PixText output crops. All outputs use the same crop coordinates and display scale; pale-blue frames identify model outputs.}
  \label{fig:ood-generation}
\end{figure}

\section{Comparison with Qwen-Image-3.0-Pro}
\label{app:qwen-api-refinement}
We compare PixText and PixText + MeanFlow with Qwen-Image-3.0-Pro on the same 1,034-image Text Refinement Bench used in the main results. The API receives each full image and all its repair targets in one bounding-box text prompt. It returns 1,027 edited images and refuses to generate edited outputs for the remaining seven inputs; their 11 repair targets count as failures in the text-accuracy metrics.

Both PixText variants achieve substantially higher text accuracy and better image preservation than the API. This shows that even an advanced image-editing model can remain unreliable for precise, localized text repair.

\begin{table}[h]
\centering
\caption{Comparison on the same 1,034 images and 2,638 repair targets. Targets in the seven cases where the API refused generation count as failures for Pass, R@1, and OCR Acc; image metrics use returned images.}
\label{tab:qwen-api-refinement}
\small
\setlength{\tabcolsep}{3pt}
\resizebox{\linewidth}{!}{%
\begin{tabular}{lrrrrrr}
\toprule
Method & Images & Pass$\uparrow$ & R@1$\uparrow$ & OCR Acc$\uparrow$ & FID$\downarrow$ & LPIPS$\downarrow$ \\
\midrule
\textbf{PixText} & 1034/1034 & 83.06\% & 92.95\% & 94.66\% & 0.34 & 0.0010 \\
PixText + MeanFlow & 1034/1034 & 81.84\% & 92.57\% & 94.58\% & 0.38 & 0.0011 \\
Qwen-Image-3.0-Pro & 1027/1034 & 28.17\% & 60.92\% & 70.13\% & 7.39 & 0.0188 \\
\bottomrule
\end{tabular}}
\end{table}

\end{document}